\documentclass[10pt,twocolumn,letterpaper]{article} 

\usepackage[numbers,sort]{natbib}
\usepackage{stfloats}
\usepackage{float}
\usepackage{subcaption}
\usepackage{pgf}

\usepackage{ICCV/iccv}
\usepackage{times}
\usepackage{epsfig}
\usepackage{graphicx}
\usepackage{amsmath}
\usepackage{amssymb}

\definecolor{myRed}{RGB}{228, 26, 28}
\definecolor{myBlue}{RGB}{55, 126, 184}
\definecolor{myGreen}{RGB}{77, 175, 74}
\definecolor{myPurple}{RGB}{152, 78, 163}
\definecolor{myOrange}{RGB}{252, 124, 4}

\usepackage{bbold}
\usepackage{amsmath,amsthm,amssymb}
\usepackage{amsfonts}
\usepackage{cases}
\usepackage{bm}
\usepackage{nicefrac}       

\usepackage{thmtools,thm-restate}

\newtheorem*{theorem*}{Theorem}

\DeclareMathOperator*{\argmax}{\arg\max}

\newcommand{\ourmethod}[0]{ECINN}

\usepackage{booktabs}
\usepackage{multirow,tabularx}
\newcolumntype{Y}{>{\centering\arraybackslash}X}
\newcolumntype{L}{>{\arraybackslash}X}

\usepackage[plain]{fancyref}

\usepackage[pagebackref=true,breaklinks=true,colorlinks,bookmarks=false]{hyperref}

\iccvfinalcopy 

\def\iccvPaperID{4348} 

\title{\ourmethod{}: Efficient Counterfactuals from Invertible Neural Networks}
\author{
Frederik Hvilsh\o j\\ 
Aarhus University, CS\\
{\tt\small fhvilshoj@cs.au.dk}
\and 
Alexandros Iosifidis\\ 
Aarhus University, ENG\\
{\tt\small ai@ece.au.dk}
\and 
Ira Assent\\ 
Aarhus University, CS\\
{\tt\small ira@cs.au.dk}
\and 
}

\begin{document}
\maketitle

\begin{abstract}
Counterfactual examples identify how inputs can be altered to change the predicted class of a classifier, thus opening up the black-box nature of, \eg, deep neural networks. 
We propose a method, \ourmethod{}, that utilizes the generative capacities of invertible neural networks for image classification to generate counterfactual examples efficiently. 
In contrast to competing methods that sometimes need a thousand evaluations or more of the classifier, \ourmethod{} has a closed-form expression and generates a counterfactual in the time of only two evaluations.
Arguably, the main challenge of generating counterfactual examples is to alter only input features that affect the predicted outcome, \ie, class-dependent features. 
Our experiments demonstrate how \ourmethod{} alters class-dependent image regions to change the perceptual and predicted class of the counterfactuals. 
Additionally, we extend \ourmethod{} to also produce heatmaps (\ourmethod{}h) for easy inspection of, \eg, pairwise class-dependent changes in the generated counterfactual examples. 
Experimentally, we find that \ourmethod{}h outperforms established methods that generate heatmap-based explanations.
\end{abstract}


\section{Introduction}\label{sec:introduction}
Deep neural networks are becoming increasingly popular and exhibit unprecedented capabilities within a range of computer vision tasks, some even surpassing human performance~\cite{noisyStudent}.
The price for such high performance is a lack of transparency.
In high stake domains like health care, autonomous transportation, or automated decision-making involving human lives, opaque models can be an issue, \eg, due to a lack of understanding of the networks.

In recent years, a great effort has been devoted to open up the black-box nature of deep neural networks for computer vision. 
Among others, heatmaps~\cite{lrp}, class-maximizing samples~\cite{simonyanVZ13}, and contrastive examples~\cite{Dhurandhar2018} have been proposed.
In this work, we mainly focus on the latter.

\begin{figure}[t!]
    \centering
    \includegraphics[width=\linewidth]{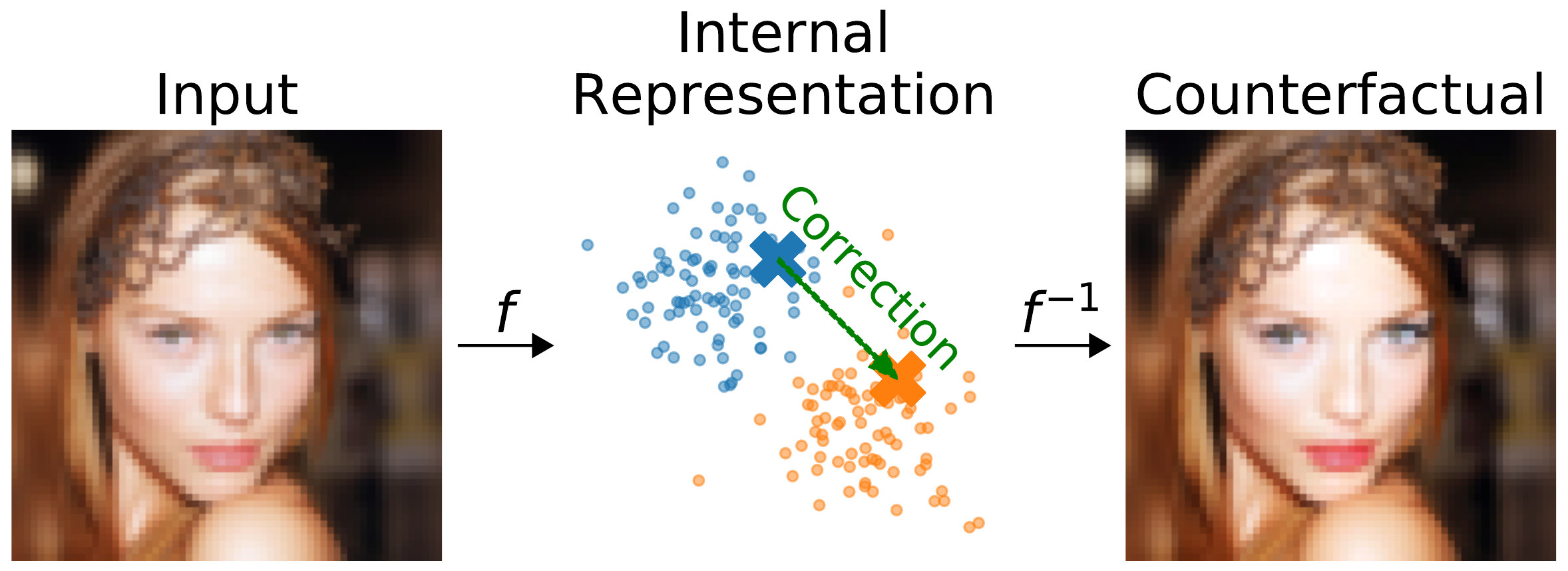}
    \caption{INN $f$ transforms image of woman \emph{without} makeup (left) into an internal representation. 
    The internal representation is corrected with closed-form expression (center). 
    Inverse INN $f^{-1}$ generates counterfactual example \emph{with} makeup (right).}
    \label{fig:frontpage}
\end{figure}

Contrastive examples are also known as counterfactual examples, even though models do not possess any causal structure as described in~\cite{pearl2015a}.
We adopt the setting from~\cite{goyal19a} and consider the generic question, ``For situation $X$, why was the outcome $Y$ and not $Z$?''  
We provide a counterfactual example to give an explanation of the form ``Had $X$ been $\hat X$, then the outcome would have been $Z$.''

Being able to provide counterfactual examples for complex neural networks has an immense potential to improve human-model-interactions.
To name but a few, surveillance systems could be assessed for biases when picking out candidates for screening, and self-driving vehicles could be better diagnosed when misinterpreting their image feeds~\cite{goyal19a}.

Good counterfactual examples are broadly agreed to be realistic, minimal, and actionable~\cite{Wachter2017,Gomez2020}. 
In the image domain, however, minimal changes are hard to measure in a semantically meaningful way. 
For example, an adversarial attack changing just one pixel can be enough to change predictions~\cite{su2019one}. 
While such an attack is minimal in terms of the number of pixels changed, it is not realistic and thus not desired in the context of counterfactual examples.
As such, we argue that the main challenge is to generate \emph{perceptible} and \emph{realistically} looking images where only class-relevant features are changed.
For example, lips, eyes, and maybe cheeks would change if makeup was applied to a face but not hair color or the background. 
Recently, many methods for generating counterfactual examples have been proposed~\cite{Akula2020, Dhurandhar2018, Pan2019, Gomez2020, Vermeire2020, cheng2021dece, Jacovi2021, Wu2021, goyal19a, VanLooveren2019}. 
A common drawback for all the methods is that they need to query the model under consideration many times.
To the best of our knowledge, we, in contrast, introduce the first algorithm that produces a counterfactual example from just one query of the model and one reverse pass.

A one-pass-solution is possible because we utilize Invertible Neural Networks (INNs)~\cite{nice}, which, in contrast to usual discriminative models, preserve all information between input and output layers and, in turn, allow recovering inputs exactly from their outputs.
Additionally, INNs are known to have semantically organized latent spaces where translations in specific directions result in semantic changes in the input space~\cite{realnvp}.
As such, it can be argued that INNs are ideal for combining generative and discriminative capabilities for neural networks~\cite{iresnet}.

We propose the method Efficient Counterfactuals from INNs (\ourmethod{}), which utilizes already trained INN classifiers to transform inputs into an internal representation.
Among internal representations, closed-form counterfactual corrections become possible.
Counterfactual examples are then generated by transforming the corrected internal representations through the reverse INN.
\Fref{fig:frontpage} depicts the high-level structure of \ourmethod{}. 
The figure shows how an input image of a woman without makeup (left) is transformed by an INN denoted $f$ into an internal representation (center).
The internal representation is then corrected, as indicated by the green arrow, before being reverted by $f^{-1}$ to form a counterfactual example that wears makeup (right). 

We demonstrate experimentally how \ourmethod{} produces counterfactual examples that change class-dependent features while class-independent features are left largely untouched. 
We further demonstrate how visualizations of discrepancies between inputs and counterfactuals outperform established heatmap generation methods which produce surprisingly noisy heatmaps on conditional INNs.
Backed by a simple experiment, we conjecture that such differences are due to the absence of ReLU activation in INNs.


\section{Related Work}\label{sec:related-work}
In this work, we utilize INNs to generate counterfactual examples. 
Therefore, we devote some attention to INNs in this section but keep the main focus on explaining neural networks with counterfactual examples. 

\subsection{Explaining Neural Networks}\label{sec:rw-counterfactual-explanations}
\paragraph{Counterfactual Examples.}
In recent years, many methods have been proposed for synthesizing counterfactual examples or identifying counterfactual features on various types of data. To name but a few, \cite{goyal19a, Akula2020, VanLooveren2019, Vermeire2020, Wang2020} operate on image data, \cite{Jacovi2021, Wu2021, SinHan2020} consider text, and yet other methods operate on relatively low dimensional data compared to images and text~\cite{Gomez2020, cheng2021dece, Wachter2017}.

Methods for generating counterfactual examples can be categorized by the insights needed into the predictive model. 
Methods from the first category consider the predictive model as opaque and need no insight.
Methods from the second category utilize gradients of the predictive model, while methods from the last category use internal data representations of the predictive model.
All methods mentioned here have~the drawback that they need to query the predictive model multiple times.
\cite{Wang2020} identifies counterfactual regions in input images but does not generate counterfactual examples. 
In~contrast, after a preprocessing step that needs to be done only once, our method uses a single forward and inverse pass through the model to generate a counterfactual example.

In the first category, methods operating on opaque models typically work by iteratively generating candidate sets of counterfactual examples and then querying the predictive model to test candidates. \cite{Gomez2020} utilizes a greedy heuristic from simple data statistics to determine what input features to perturb, while~\cite{Sharma2019} uses a genetic algorithm.
\cite{Vermeire2020} segments input images into super-pixels and use a greedy algorithm to perturb super-pixels to identify which regions affect the output of the classifier.
On text data, \cite{Wu2021} finetunes a GPT-2 model~\cite{radford2019language} to generate similar sentences to the input sentence to generate new candidates. 
Similarly, \cite{VanLooveren2019} uses autoencoders and KD-trees to identify images similar to input images to speed up the search for candidates that change the prediction of the predictive network.
In comparison to a forward and inverse pass which \ourmethod{} uses, the default maximum queries of the classifier in the official code of~\cite{VanLooveren2019} is a thousand. 

The second category of methods employs gradient optimization techniques to identify inputs that change the decision of the predictive model.
We note that such ideas are not new. Previous work, albeit from a different perspective, has developed methods for synthesizing inputs that maximize desired (output) neurons of a given network. 
For example, \cite{simonyanVZ13} uses gradient descent with an $L_2$-norm prior loss on a random input to maximize output neurons. 
\cite{amax-fooled} includes a local pixel variation prior in the loss to obtain more realistically looking features in the generated images. 
Even though the methods give insights into the inner workings of the classifier, they suffer from generating unrealistic images.
More recently, \cite{Pan2019} proposed to train a generative model to, given the input image, make new alternative images that would change the prediction of the classifier.
In a similar vein, \cite{Dhurandhar2018} utilizes a pretrained and fixed autoencoder to identify a latent code that generates the desired output through gradient optimization. 

The third category of methods contains two different strategies. First, \cite{goyal19a} considers convolutional neural networks as a composition of a (convolutional) feature extractor and a classification network and proposes two algorithms to mix fibers of 
the feature extractor applied to the input and a sample from the counterfactual class. 
Second, \cite{Akula2020} similarly uses a part of the classifying network as a feature extractor to cluster such features. 
The result is an identification of semantic features like stripes, wool, etc. 
A gradient descent algorithm then learns how to add or remove from an input to obtain a counterfactual example.

The work we present in this paper fits best into the third category. 
However, our approach is conceptually different.
Instead of generating counterfactual examples from an ``arbitrary'' neural network, we choose a specific family of neural networks, INNs, to generate counterfactual examples efficiently without the use of multiple queries of the model or gradient computations. 

\paragraph{Heatmaps.}
There exist many methods that produce explanations in the form of heatmaps.
Some methods work on black-box predictive models~\cite{shap, lime, Zintgraf2017, Chang2019} while others utilize gradient-like computations on the predictive model~\cite{lrp, gbp, deeplift, intgrad, deconvnet, unifiedShap}.
In this work, we compare our method against four methods from the latter category. 
i) DeepLift~\cite{deeplift} which is based on discrepancies between modified gradients of the input and a non-informative reference point, ii) Integrated Gradients~\cite{intgrad} (IntGrad) which approximates integrals of gradients from a reference point to the input, iii) GradSHAP, and iv) DeepLiftSHAP which are two related methods for approximating SHAP values~\cite{unifiedShap}. 
We refer the reader to~\cite{unifiedShap} for a detailed description of the four methods.

\subsection{INNs as Generative Classifiers}\label{sec:rw-normalizing-flows}
INNs have gained wide attention as unsupervised generative models which allow generating realistically looking ``fake'' samples~\cite{nice, realnvp, glow}; when used for generative modeling, INNs are typically referred to as Normalizing Flows.
Despite hidden in appendices, both~\cite{realnvp} and \cite{glow} present samples generated from class-conditional INNs. 
Later, it was explicitly described how to follow the INNs by a Gaussian mixture model (GMM) to obtain a generative classifier~\cite{Nguyen2019, gmmflow}, which both allows class-conditional sampling and sample classification.
However, adding classification abilities comes at a price.
As demonstrated in \cite{ibinn}, there is a trade-off between classification performance and the quality of the generated fake images.
The work introduces an information bottleneck loss, which explicitly trades off the classification and generation performance through a hyperparameter $\beta$.
\cite{ibinn} further introduces a new invertible model architecture, which we refer to as IB-INN.

Regarding interpretability, \cite{expl-inn} shows how conditional INNs can be trustworthy classifiers by visualizing decision spaces, comparing class similarities, and computing posterior heatmaps.
In this work, we further show conditional INNs to be trustworthy classifiers by using them for generating counterfactual examples. 


\section{Efficient Counterfactual Examples}\label{sec:efficient-counterfactuals}
This section constitutes our main contribution.
We combine theoretical insights and practical observations from INNs to generate counterfactual examples efficiently.

\subsection{Problem Statement}\label{subsec:problem-statement}
As mentioned, counterfactual examples are samples that indicate why an input instance was predicted to be one class rather than another. 
Specifically, we modify the definition from \cite{Wachter2017} which states that counterfactual examples are statements taking the form: 
``Score $p$ was returned because variables $V$ had values $(v_1, v_2, \dots)$ associated with them. If $V$ instead had values $(v_1', v_2', \dots)$, and all other variables had remained constant, score $p'$ would have been returned.''
In the context of image classification, we define counterfactual examples as visualizations showing how the input image can be altered to change the predicted class. 
 
\paragraph{Desiderata.}
In line with the desiderata of \cite{Gomez2020} and \cite{Wachter2017}, we find that three properties are of high importance for counterfactuals to be useful.
i) \emph{Only semantically relevant features should be changed}.
For example, facial features like lips, cheeks, and eyes might change while background and hair should not when a counterfactual is generated for a face without makeup. 
ii) \emph{Counterfactuals should look realistic}.
Examples of unrealistic counterfactuals could be misplaced eyes on a face, extreme color values, or a ``one-pixel-change'' like the adversarial examples presented in \cite{su2019one}.
iii) \emph{Both tipping-point counterfactuals and convincing counterfactuals should be prioritized}.
We refer to counterfactuals on the decision boundary between the input and the target class as tipping-point counterfactuals. 
Likewise, counterfactuals, where the target class is predicted with high confidence, are referred to as convincing counterfactuals. 
Tipping-point counterfactuals are essential because they identify a minimal correction to the input.
However, they might not always make sense due to visual class differences.
For example, when changing the predicted class of a cat to a dog, a tipping-point counterfactual might fail to show how the ears should be pointy instead of hanging because the tipping-point would represent something in between.
On the contrary, a convincing counterfactual would successfully show such transformation, but potentially with too pronounced changes. 
Providing both types of explanations thus give a deeper insight into the decisions of the classifier.

We emphasize that the counterfactual examples discussed in this work are not causal as counterfactual examples described in, \eg, \cite{pearl2015a}. 
Although the ambiguity of the name is unfortunate, we stick to the naming convention to be consistent with related work.

\subsection{Conditional INNs}\label{subsec:ibinn}
We find INNs to be well suited for the counterfactual problem because they are bijective, \ie, every latent vector corresponds to exactly one input.
In contrast, typical classification models are inherently surjective, \ie, there exist many inputs which produce each output.
Identifying the best input from an output thus becomes simpler for INNs. 

It is also known that well-trained INNs have semantically organized latent spaces~\cite{realnvp}. 
We believe that when many latent representations of samples from the same class are averaged, then class-independent information like background and object orientation will cancel out and leave just class-dependent information. 
\ourmethod{} isolates such latent class-dependent information and uses it to correct latent space embeddings to generate counterfactual examples. 

A conditional INN $f$ is typically trained by computing latent vectors $z=f(X)$ from input vectors $X$ and using the latent vectors to fit a GMM to class labels $Y$.
However, to use $Z$ rather than $X$ in the GMM, one must use the change-of-variables formula, which states that
\begin{equation}\label{eq:change-of-variable}
\log p_X(x | y) = \log p_Z(f(x) | y) + \log \left| det\left( J\right)\right|.
\end{equation} 
That is, the class-conditional log density of an input $x$ in the image space, $p_X(x |y)$, is equal to the class-conditional log density of $f(x)$ in the latent space $p_Z(f(x)|y)$, but with an additional Jacobian term, $J =  \frac{\partial f(x)}{\partial x}$. 
Typically, the class-dependent latent densities are chosen to be Gaussians, $p_Z(z | y) = \mathcal{N}(\mu_y, \mathbb{1})$. 
By Bayes' rule, we notice that under a uniform prior distribution over labels, $p(y) = 1/K$ for $K$ classes, the log posterior probability becomes
\begin{equation}\label{eq:propto}
    \log p_X(y | x) = \log\frac{p_X(x|y)}{\sum_{y'}p_X(x|y')} \propto -||f(x) - \mu_y||^2.
\end{equation}
From \Fref{eq:propto}, we see that independent of the Jacobian determinant, latent vector $z=f(x)$ will be predicted to be from the class $y$ with the closest model mean, $\mu_y$. 
In turn, the latent space of the classifier can be analyzed under $L_2$-norms instead of less efficient and complex densities $p_X(x|y)$, which depend on the Jacobian determinant. 
In the following subsection, we present how \ourmethod{} utilizes this insight to produce counterfactual examples efficiently.

\begin{figure}[b]
    \centering
    \includegraphics[width=0.97\linewidth, page=2]{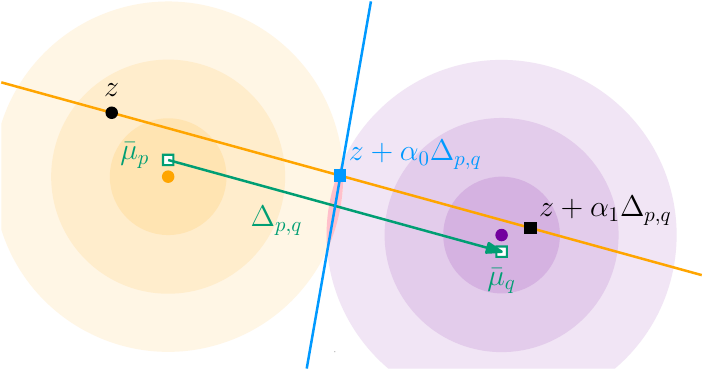}
    \caption{Latent space corrections by \ourmethod{}.}
    \label{fig:intersecting-hyperplanes}
\end{figure}

\subsection{\ourmethod{} }\label{sec:generating-examples}
At a high level, \ourmethod{} transforms images into a latent space through an INN $f$.
In the latent space, a closed-form expression is used to correct the latent embedding to change the predicted class of the INN. 
From the corrected embedding, a counterfactual is generated by the inverse INN $f^{-1}$.

As a preprocessing step that needs to be done only once, we group the training samples by their classified output, $G_j = \{x  | C(x) = j\}$, where $C(x) = \argmax_{y}p_X(y|x)$ is the predicted class.
Afterwards, we compute mean latent vectors $\bar\mu_j = \frac{1}{|G_j|}\sum_{x\in G_j}f(x)$ for each class $j$ and define the vector from $\hat\mu_p$ to $\hat\mu_q$ as $\Delta_{p,q} = \bar\mu_q - \bar\mu_p$.

Given a target class $q$ and an input $x$, a counterfactual example $\hat x^{(q)}$ is produced from the predicted class $C(x) = p$ by adding a scaled version of $\Delta_{p, q}$ to the latent space embedding $z=f(x)$ and inverting it through the INN,
\begin{equation}\label{eq:counterfactual}
    \hat x^{(q)} = f^{-1}( f(x) + \alpha \Delta_{p, q} ).
\end{equation}
As indicated by \Fref{eq:counterfactual}, generating a single counterfactual example requires just one evaluation of $f$ and $f^{-1}$.

To follow our third desideratum and provide both tipping-point and convincing counterfactuals, we compute two counterfactuals for each input with different values of $\alpha$.
First, we choose $\alpha_0$ to produce a tipping-point counterfactual, which potentially reveals minimal semantic changes in the image space to change the predicted class.
In the latent space, $\alpha_0$ will be the value that moves the latent vector exactly onto the decision boundary between the input and target class.
Due to \Fref{eq:propto}, $\alpha_0$ is identified analytically such that $||z+\alpha_0\Delta_{p,q} - \mu_p|| = ||z + \alpha_0\Delta_{p,q} - \mu_q||$.
The closed-form expression for $\alpha_0$ is given in the supplementary material along with a proof.
Second, we choose $\alpha_1$ such that the target class $q$ is predicted with high confidence to produce a convincing counterfactual.
$\alpha_1$ is chosen heuristically to be $\alpha_1 = \frac{4}{5} + \frac{\alpha_0}{2}$. 
Although it is not guaranteed that the counterfactual example generated is predicted to be from the target class, \ie, $C(\hat x^{(q)}) = q$, we observed that the relation holds in practice.

In \Fref{fig:intersecting-hyperplanes}, we illustrate the intuition of our method.
The figure shows two unit variance normal distributions in the latent space. 
The blue line indicates the decision boundary between the two normal distributions, and the orange line is the line that passes through $z$ in direction $\Delta_{p,q}$.
With green squares, we indicate the two computed means $\bar\mu_p$ and $\bar\mu_q$, that are used to define $\Delta_{p,q}$ (green arrow).  
The two points of interest are the blue square on the intersection of the blue and the orange line and the black square to the right.
According to the model, the blue square is equally likely to stem from either of the two classes, and the black square is very likely to stem from class $q$.
In the experimental section, we visualize both points as counterfactual examples in the image space by inverting them through $f^{-1}$.

With \ourmethod{}, we have connected the well-suited properties of INNs to the observation that the latent space of the conditional INNs is easy to analyze due to the relation shown in \Fref{eq:propto}.  

\paragraph{\ourmethod{}h.}
A common assumption for heatmap generation methods is that removing information from pixels identified as important should cause the predicted class to become less likely under the model~\cite{Samek2017}. 
Following this, we introduce \ourmethod{}h for producing pairwise heatmaps by highlighting pixel discrepancies between inputs, $x$ and the generated counterfactual examples, $\hat x^{(q)}$.

\begin{equation}\label{eq:heatmap}
    h(x, q) = \hat x^{(q)} - x.
\end{equation}

\Fref{eq:heatmap} yields an estimate of how important each pixel of the input is for expressing class $p$ of the compared to a target class $q$.
A large absolute value means that the associated pixel needs to change a lot for the predicted class to change, \ie, it intuitively contains a high amount of class-dependent information.
In turn, such pixel can be interpreted as having a large impact on the predicted class. 
On the contrary, values close to zero indicate that associated pixels can remain unchanged while the predicted class change.
Such pixels have no class-dependent information to remove and will probably not change the predicted class if altered.

The two heatmap generation methods DeConvNet~\cite{deconvnet} and GuidedBackProp~\cite{gbp} are examples of how ReLU-activations have a large effect on gradient-based heatmaps. 
Both methods demonstrate how applying ReLUs to gradient computations removes noise in the generated heatmaps.
Because INNs need to be invertible, components like ReLU activations are not directly applicable.
Therefore, we expect established heatmap generation methods introduced for networks with ReLU activations to be underperforming on conditional INNs.
In contrast, as heatmaps generated by \ourmethod{}h were explicitly designed for INNs and are not based on gradients, we expect them to be less noisy and of a higher quality.

In conclusion, we introduce \ourmethod{} which allows computing counterfactuals efficiently by utilizing properties of INNs.
\ourmethod{} complies with our first two desiderata by using INNs to generate counterfactuals from latent space directions, which represent class-dependents changes while leaving out most class-independent information.
Furthermore, by providing both tipping-point and convincing counterfactuals, we follow the third desideratum.
Finally, we present the \ourmethod{}h extension, which generates heatmaps that allow easy inspection of class-dependent changes in the generated explanations. 


\section{Experiments}\label{sec:experiments}
In this section, we evaluate how our counterfactual examples perform.
Our experiments show how \ourmethod{} produces meaningful counterfactual examples across three different image datasets, changes class-dependent features while maintaining class-independent features, and outperforms established heatmapping methods.

\paragraph{Experimental Details.} 

We evaluate \ourmethod{} on a synthetic FakeMNIST dataset, on the MNIST dataset~\cite{mnist}, and the CelebA-HQ dataset~\cite{celeba}. 
On all three datasets, classification errors of the IB-INN models are comparable to those of a standard classification network (see \Fref{tab:stats} in the supplementary material).
For all our experiments, we have trained IB-INN models ``as-is.''\footnote{We adopted models and training code from \url{https://github.com/VLL-HD/IB-INN}.}
We note that the $\beta$-value of the IB-INN loss influences the performance of our method.
In the presented experiments, we found that values close to one strike a good balance between classification accuracy and generative performance.
In the supplementary material, we provide an overview of all models used, their hyperparameters, and their performances.
We also include additional samples of all plots. 
Results presented in this section are all with samples from the test set and were found to be consistent across samples.

We provide code in an iPython Notebook, \verb!code.ipynb!, which can be uploaded to Google Colab and run with one run command.
Upon submission, we plan to release our code.
Finally, we suggest reading this section on a screen to enable zooming on the figures.

\begin{figure}[b]
    \centering
    \includegraphics[width=0.78\linewidth]{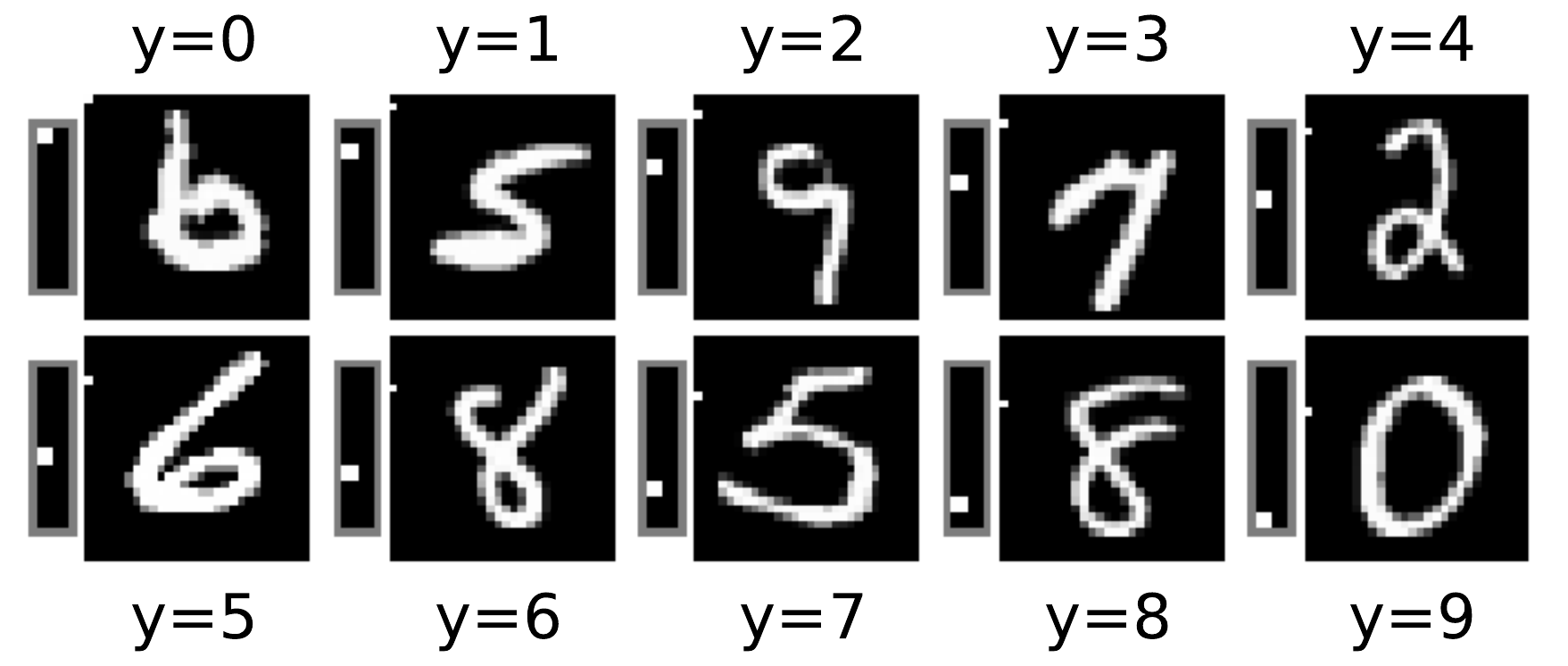}
    \caption{
        Random samples from the FakeMNIST dataset.
        For improved readability, smaller rectangles to the left of images magnify the top left $10\times2$ pixels, indicating the class.
    }
    \label{fig:fakemnist-samples}
\end{figure}
\subsection{FakeMNIST}

The goal of the first experiment is to verify \ourmethod{} in a controlled setting. 
We construct an image dataset where less than two percent of the pixels are class-\emph{dependent}.
The remaining pixels are \emph{independent} of the class label.
As argued, a proper counterfactual example for a well-trained model should alter only the class-dependent pixels. 
Additionally, if the class-dependent pixels are not present, such an instance should be equally likely to be from any class.

We generate a dataset by computing new uniformly random labels for all MNIST samples and color the top left $10\times1$ pixels accordingly. 
For example, if an image gets assigned label ``5,''  we color the sixth pixel in the left column white.
As such, the labels are independent of the depicted digits and only depend on the top left pixels.
\Fref{fig:fakemnist-samples} shows a sample from each of the ten classes.  
The top-left pixels vary with the labels ($y$), and the depicted digits do not.

\begin{figure*}
    \centering
    \includegraphics[width=0.65\linewidth]{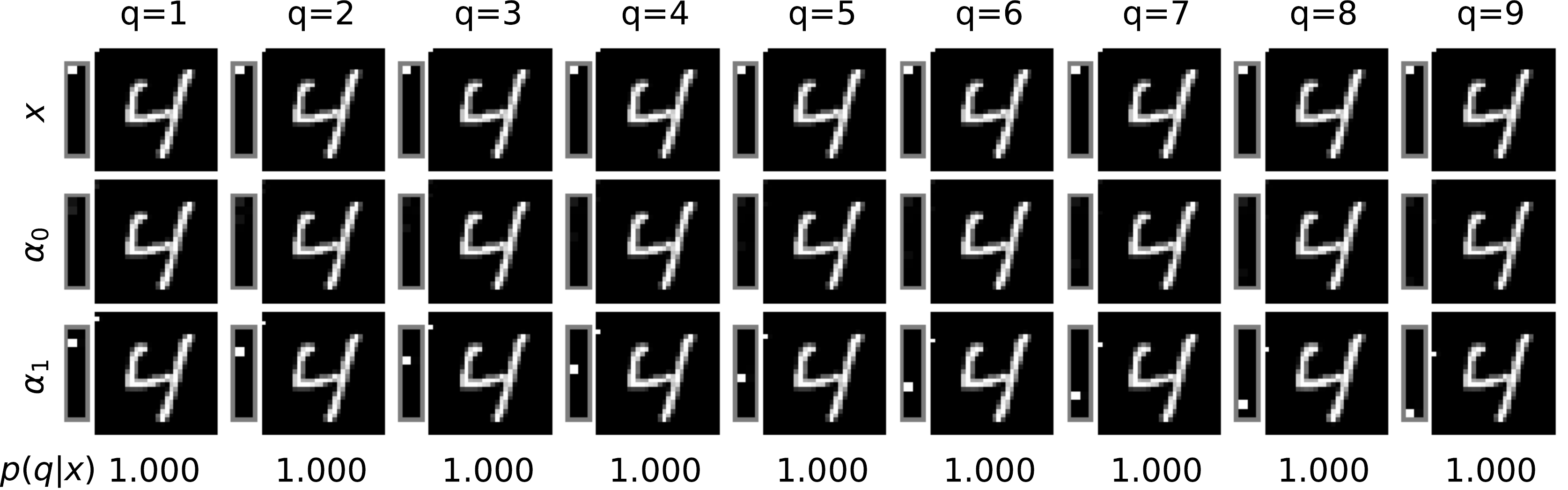}
    \caption{Counterfactual examples generated for the FakeMNIST dataset. Columns represent targets $q$ and rows are input, $\alpha_0$ counterfactuals, and $\alpha_1$ counterfactuals, respectively.}
    \label{fig:fake-counterfactuals}
\end{figure*}

In \Fref{fig:fake-counterfactuals}, we have drawn a random sample from the class $y=0$ (first row) and display a counterfactual example for $\alpha_0$ (second row), which is equally likely to stem from the class $y=0$ and the target class $q$. 
Additionally, the figure includes a counterfactual example for $\alpha_1$, which represents a high confidence ($p(q|x)$ close to 1) of the classifier (third row). 
Each column corresponds to a different target class $q$ as indicated by the labels above each column.

\Fref{fig:fake-counterfactuals} shows that the dot in the top left corner of the input does change position while the class-independent digit remains unchanged as expected.
Specifically, the third row from left to right reveals how the dot in the top left corner travels downwards to end in the tenth pixel.
Notably, the second row has almost no dot, which aligns well with the interpretation about equally likely class probabilities above.

\subsection{MNIST}
\begin{figure}[b!]
    \centering
    \includegraphics[width=\linewidth]{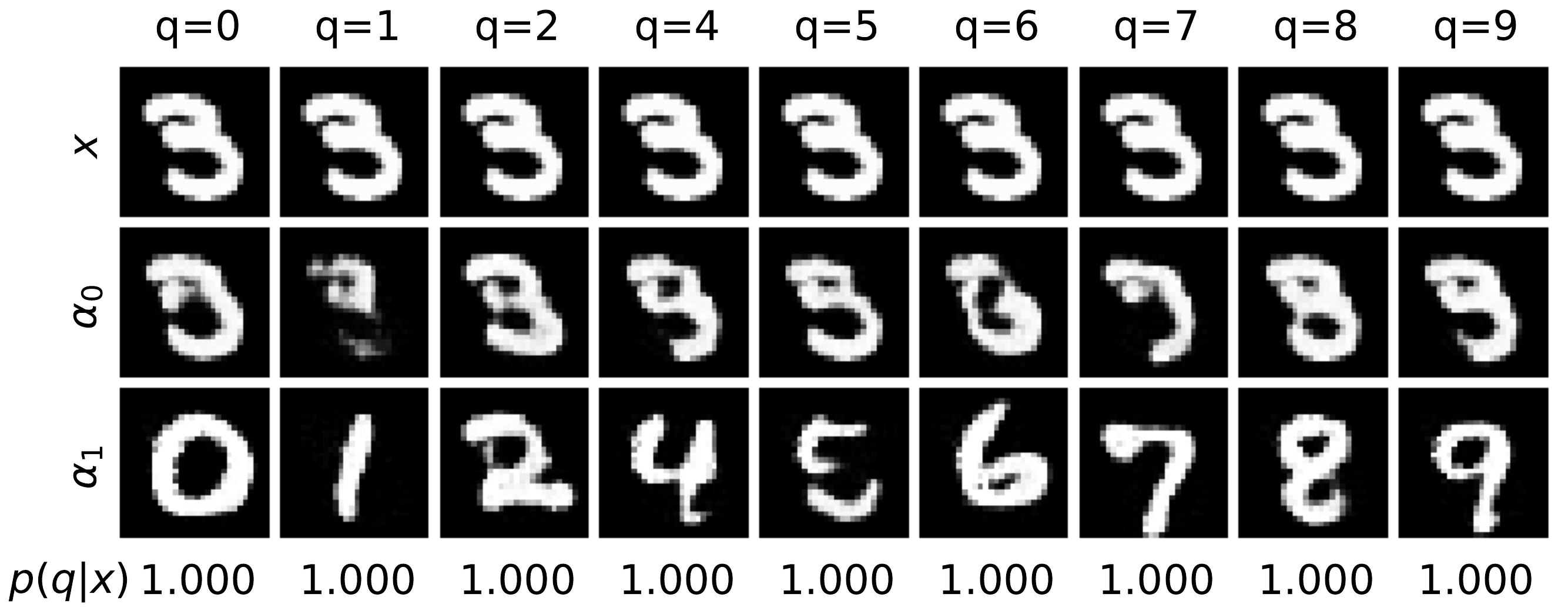}
    \caption{Counterfactual examples for a sample of a three with different target classes, $q$.}
    \label{fig:mnist-multi-target}
\end{figure}

Next, we apply \ourmethod{} to the MNIST dataset. 
We seek to investigate two properties.
First, we verify our second desideratum, \ie, that \ourmethod{} produces realistic counterfactual examples.
Second, we investigate how well class-independent features like font-weight, tilt, and size are maintained by \ourmethod{}, \ie, our first desideratum.

\paragraph{Realistic Counterfactuals.}
In \Fref{fig:mnist-multi-target}, we depict counterfactual examples in the same fasion as \Fref{fig:fake-counterfactuals}. 
The figure shows how an image of a three is properly transformed into any of the remaining nine classes.
Note that in the second row, the counterfactual examples are in many cases such that even a human might mistake the image for both the input and target class. 
By contrast, the third row contains samples where the three has successfully transformed into the target class.
This experiment demonstrates that \ourmethod{} complies with our second desideratum by generating realistic counterfactuals.

\begin{figure}[b]
    \centering
    \includegraphics[width=\linewidth]{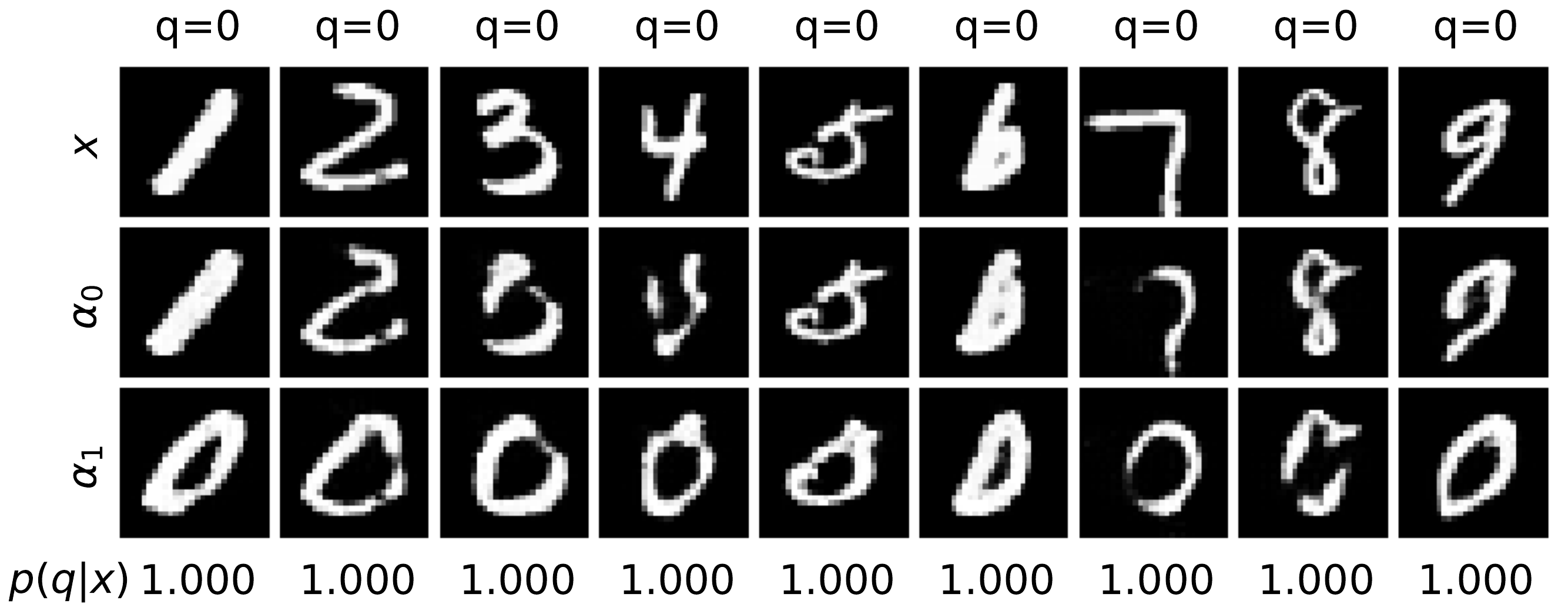}
    \caption{Counterfactual examples with $q=0$.}
    \label{fig:mnist-all-to-one}
\end{figure}
\begin{figure}[b]
    \centering
    \includegraphics[width=\linewidth]{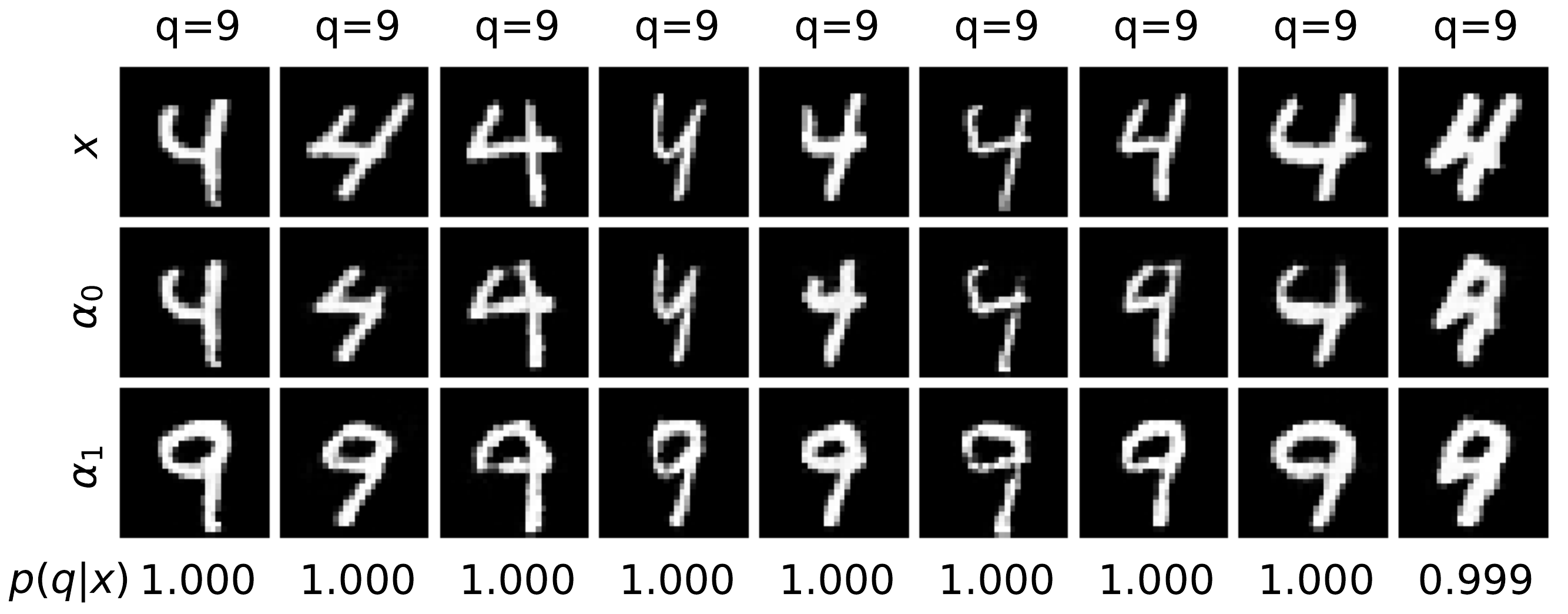}
    \caption{Diverse counterfactuals for same input and target class.}
    \label{fig:mnist-style}
\end{figure}
\begin{figure*}[t]
    \centering
    \includegraphics[width=\textwidth]{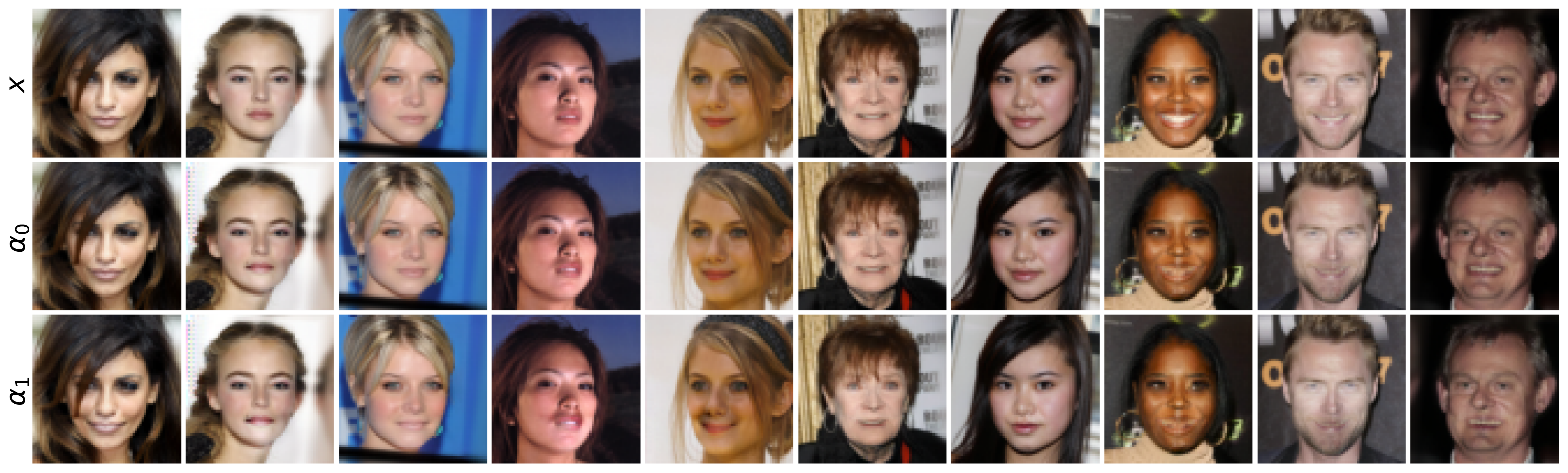}
    \caption{
        Counterfactual examples for frowning and smiling faces. 
        First row is the input. Second and third row are generated with $\alpha_0$ and $\alpha_1$, respectively. First five columns have target $q=\text{smile}$ and last five columns $q=\text{frown}$. $p(q|x) > 1-10^{-4}$ for all samples.
    }
    \label{fig:celeba}
\end{figure*}

\paragraph{Class-Independent Properties.}
In \Fref{fig:mnist-all-to-one} and \ref{fig:mnist-style}, we demonstrate how class-independent properties like font-weight, tilt, and size are preserved during counterfactual generation. 
First, \Fref{fig:mnist-all-to-one} includes nine different inputs (first row), each from a different class, that are all translated to the target class, $q=0$.
We observe that the nine outcomes (row three) are perceptually different while resembling the target class.
Each counterfactual example maintains class-independent properties from the input while resembling the target class.
For example, the narrow and tilted one (first column) becomes a narrow and tilted zero.
Similarly, explaining the pointy five yields a pointy zero.
The observations suggest that \ourmethod{} maintains properties that are not directly dependent on the label. 

In \Fref{fig:mnist-style}, we further investigate how class-independent properties of the input images are maintained.
We sample nine different images from the class $y=4$ and compute their counterfactual examples for the target class $q=9$.
We observe how bold inputs yield bold counterfactuals; likewise, slim inputs yield slim counterfactuals.
Similar observations can be made for, \eg, tilt, size, and shapes.

In conclusion, we observe that for the MNIST dataset, \ourmethod{} produces counterfactual examples which comply with our desiderata by realistically changing both the predicted and the perceived class while maintaining class-independent features such as font-weight, tilt, and size.

\subsection{CelebA-HQ.}
To evaluate \ourmethod{} on a more diverse and complex dataset, we extend our experiments to the CelebA-HQ dataset. 
We train IB-INNs to predict various labels, where each label occurs in at least $45\%$ of the dataset.

\paragraph{Counterfactual Examples.}
In \Fref{fig:celeba}, we show counterfactual examples as for MNIST, but on the smile versus frown label; similar plots for other labels can be found in the supplementary material.
The first five columns depict how \ourmethod{} turns frowning people into smiling ones, while the last five columns make smiling people frown.
First, we observe that class irrelevant features such as hair, skin color, and backgrounds remain perceptually unchanged as desired.
Second, we notice that some of the counterfactual examples in the last row look unrealistic. 
In particular, it seems to be hard for the method to open and close mouths. 
In some cases, we also observe small artifacts like the ones in the left-most pixels of the second column. 
Based on our MNIST experiments, which did not suffer from computational limitations, we believe that scaling from roughly $40$ million parameters that our models use to around $200$ million parameters (as is common with previous work~\cite{glow}) can remove the artifacts and generate higher quality counterfactual examples. 
Furthermore, the low-resolution version of CelebA-HQ that we use due to limited resources is arguably harder to synthesize than higher resolutions.

\paragraph{\ourmethod{}h.}

\begin{figure*}[t!]
    \begin{subfigure}{0.580\textwidth}
        \centering
        \includegraphics[width=\linewidth]{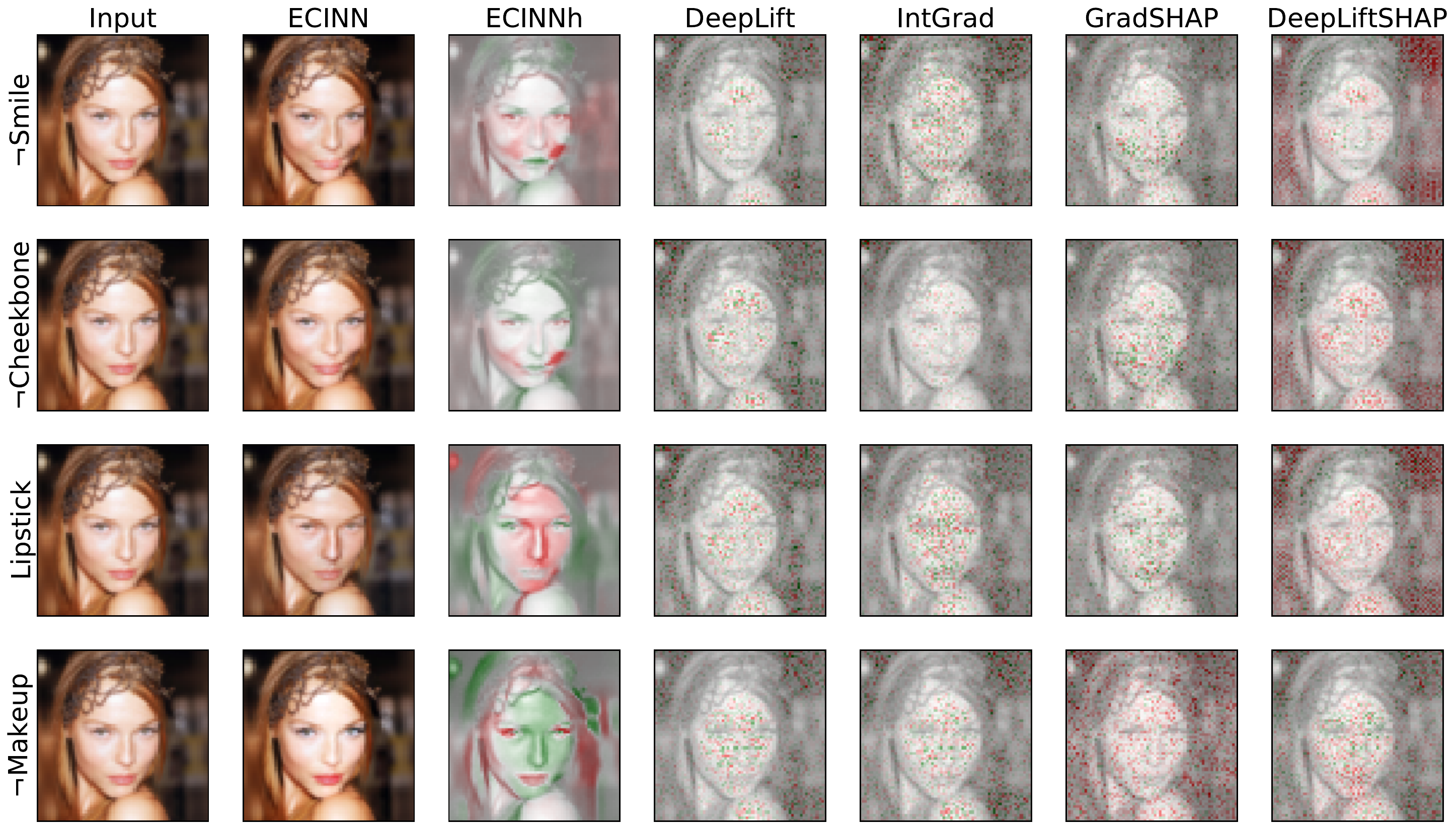}
        \caption{Predictive models: IB-INN~\cite{ibinn}}
        \label{fig:ibinn-heatmaps}
    \end{subfigure}
    \begin{subfigure}{0.410\textwidth}
        \centering
        \includegraphics[width=\linewidth]{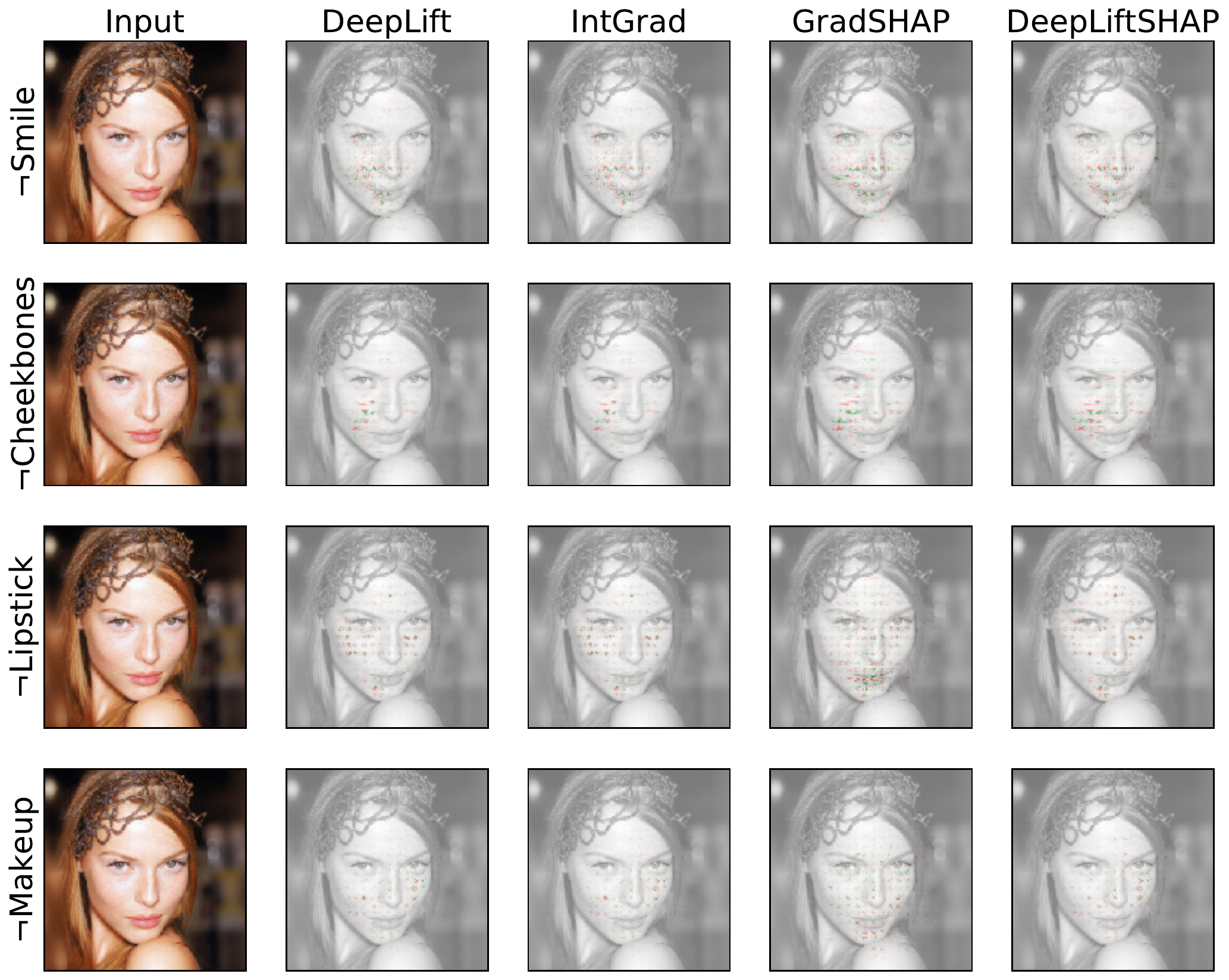}
        \caption{Predictive models: InceptionV3~\cite{inceptionv3}}
        \label{fig:inception-heatmaps}
    \end{subfigure}
    \caption{Comparison of heatmaps on INNs. Rows represent identical model architectures trained on different labels. Columns represent methods. Text on the left indicates predicted label.}
    \label{fig:heatmaps}
\end{figure*}

From the above experiments, we observe that \ourmethod{} can identify the image locations connected to the class of interest. 
Identifying such locations has been the main focus of various heatmap generation methods focusing on the explainability of deep learning models. 
To demonstrate this capability, we compare \ourmethod{}h to Integrated Gradients~\cite{intgrad}, DeepLift~\cite{deeplift}, DeepLiftSHAP~\cite{unifiedShap}, and GradSHAP~\cite{unifiedShap}.\footnote{All methods implemented through the PyTorch Captum framework, \url{https://github.com/pytorch/captum}.}
We train four models to predict whether persons are smiling, have high cheekbones, wear lipstick, or wear heavy makeup.  
Since the labels are binary, we can directly compare the methods because the target class $q$ is defined by the predicted class $p$, \ie, $q = 1-p$. 

\Fref{fig:ibinn-heatmaps} depicts heatmaps made by \ourmethod{}h (third column) and by the four mentioned methods (last four columns). 
The figure further includes the input image and the counterfactual examples generated by \ourmethod{} (first two columns, respectively).
The four different rows resemble the four IB-INNs trained on the different labels.
The text to the left indicates what the classifier has predicted the input to be (the symbol $\neg$ means ``not''). 
Comparing the heatmaps across models and methods, we observe that \ourmethod{}h consistently produces more coherent and meaningful heatmaps.
For example, in the first row, the heatmap shows how \ourmethod{} puts a large emphasis on the lips, cheekbones, and eyes, which are all associated with smiling. 
In contrast, the four other heatmaps are more scattered over the whole input and only slightly denser in the facial regions.

An additional property of \ourmethod{}h is that it gives a rich insight into the behavior of the model.
Inspecting the third row in \Fref{fig:ibinn-heatmaps} reveals that when predicting the lipstick label, the model puts emphasis on the whole face and not just the lips to classify the sample.
As such, \ourmethod{}h reveals how the model has effectively learned to detect makeup as a whole rather than lipstick.
Further comparing our heatmap of the lipstick model to that of the makeup model confirms this insight. 
Up to a sign, the two heatmaps are almost identical.
In turn, the two models emphasize very similar features to predict lipstick and makeup, respectively. 
A similar observation can be made between smiles and high cheekbones. 
As the lipstick and makeup labels are almost certain to correlate, the behavior we observe is expected.
The important thing to notice is that our method, in contrast to the others, identifies this behavior.
For INNs, we conclude that heatmaps computed with \ourmethod{}h are of higher quality than competing methods. 

\paragraph{Model Architectures.}
Even though it is well known that Integrated Gradients can be noisy, we find the results in \Fref{fig:ibinn-heatmaps} to be unusually noisy. 
We hypothesize that the reason is the absence of ReLU activations of INN layers.\footnote{INNs do have ReLUs but only in auxiliary networks of coupling layers~\cite{realnvp}, where they do not help filter noise.}
As described above, there are no ReLU activations in INNs to filter out noise during backpropagation.
This might explain why especially Integrated Gradients work well on, \eg, the Inception-V3 network~\cite{inceptionv3}, but not on INNs.
To investigate our hypothesis, we train Inception-V3 networks to classify the same samples as the IB-INNs, but in a $224\times224$ resolution due to the architecture of Inception-V3.
The resulting heatmaps are depicted in \Fref{fig:inception-heatmaps}. 
Comparing the four methods to \Fref{fig:ibinn-heatmaps}, they are less noisy and even to some extent coincide with the areas of the heatmaps of \ourmethod{}h. 
We here compare heatmaps across different datasets and architectures and cannot conclude the definitive cause.
However, this observation supports our hypothesis about noisy explanations under the absence of ReLU activations.

In summary, our experiments demonstrate that \ourmethod{} changes class-dependent features such as the shape of digits or the expression of a smile while
leaving leaves class-independent features like tilt, font-weight, and background largely untouched. 
The experiments further highlight how heatmaps generated by \ourmethod{}h gives more faithful and coherent explanations than the methods we compare against. 

\section{Conclusion}
We introduce \ourmethod{} as an efficient method for computing counterfactual examples, requiring only a forward and an inverse pass.
\ourmethod{} transforms input images into a latent space where counterfactual corrections of latent vectors have a closed-form expression.
Through an inverse INN, counterfactual examples are generated from the corrected latent vectors.
In compliance with our desiderata, \ourmethod{} generates counterfactual explanations that i) change only class-dependent features, ii) are realistic, and iii) are diverse.
These properties of ECINN are further employed in the proposed ECINNh to produce high-quality heatmaps for conditional INNs, while heatmaps of established methods are noisy. 

\clearpage

{\small  
    \bibliographystyle{ICCV/ieee_fullname}
    \bibliography{main.bib}
}

\clearpage

\appendix
\section{Analytical $\alpha$-value, $\alpha_0$} 
Define $y(\alpha) = z + \alpha \Delta_{p, q}$ to be the line intersecting $z$ with direction $\Delta_{p, q}$. 
We wish to identify the intersection between $y(\alpha)$ and the hyperplane that constitutes the decision boundary between the two normal distributions $\mathcal{N}(\mu_p, \mathbb{1})$ and $\mathcal{N}(\mu_q, \mathbb{1})$.
Due to the simplicity of the covariance matrices of the normal distributions, we can define $w = \mu_q - \mu_p$ and $b = -\left(\frac{\mu_p + \mu_q}{2}\right)^\intercal w$ to form the decision boundary
\begin{equation}
    \label{eq:decisionboundary}
    w^\intercal x + b = 0.
\end{equation}
\Fref{eq:decisionboundary} corresponds to the blue line in \Fref{fig:intersecting-hyperplanes}.

To find the $\alpha$-value which corresponds to the intersection, set $x  = z + \alpha \Delta_{p, q}$ and solve for $\alpha$ in \Fref{eq:decisionboundary}:
\begin{align}
    && w^\intercal ( z + \alpha \Delta_{p, q} ) + b &= 0\\
    \Rightarrow && \alpha w^\intercal \Delta_{p, q} &= -( w^\intercal z + b )\\
    \Rightarrow && \alpha &= -\frac{w^\intercal z + b }{w^\intercal \Delta_{p, q}}.
\end{align}

\section{Experimental Details} \label{sec:details}
In \Fref{tab:stats}, we provide an overview of hyperparameters and performances of the networks used in this work. 

\paragraph{IB-INN.}
We have trained IB-INN models ``as-is''\footnote{IB-INN code: \url{https://github.com/VLL-HD/IB-INN}} and adjusted only the $\beta$-value of the loss function.
On FakeMNIST and MNIST, the IB-INN models were trained for 60 epochs with stochastic gradient descent and a milestone scheduler stepping from learning rate $0.07$ to $0.007$ after 50 epochs. 
On CelebA-HQ, the IB-INN models were trained for 800 epochs with the Adam optimizer~\cite{adam} and a milestone scheduler stepping with a factor $\frac{1}{10}$ after every 200 epochs. 

\paragraph{Inception-V3.}
In the last subsection of the main paper, we compare the heatmaps of conditional INNs to heatmaps of the Inception-V3~\cite{inceptionv3}.

We used the inception network ``as-is''\footnote{Inception-V3 code: \url{https://pytorch.org/vision/stable/models.html\#inception-v3} }
with the exception that we turned off the auxiliary classifier and changed the output layer to have only two output neurons.
We trained the models with default parameters of the Adam optimizer and learning rate $0.001$ for 9 epochs. 
After 9 epochs, the models started overfitting.
We did not optimize the learning rate or other hyperparameters.

\begin{table}[t]
    \centering
    \begin{tabularx}{\linewidth}{p{0.26\linewidth}cccc}
    \toprule
      Model                 & \multicolumn{3}{c}{IB-INN}    & Inception-V3\\
                            & $\beta$   & BPD   &  Err.     & Err.\\
     \midrule
     FakeMNIST              & $1.4$*  & $1.77$  & $0\%$   & - \\
     MNIST                  & $1.4$*  & $1.89$  & $0.85\%$   & - \\
     \midrule
      \multicolumn{5}{c}{CelebA-HQ}\\
        Smile           & 1 & 3.32 & 7.42\%  & 6.62\%  \\
        High cheekbones & 1 & 3.09 & 14.38\% & 13.74\% \\
        Lipstick        & 1 & 3.06 & 4.87\%  & 5.70\%  \\
        Heavy makeup    & 1 & 3.08 & 12.68\% & 10.84\% \\
     \bottomrule
    \end{tabularx}
    \caption{Hyperparameters, negative log-likelihood measured in bits per dimension (BPD), and error rates for the models used in this work. *$1.4$ was rounded from $1.4265$.}
    \label{tab:stats}
\end{table}

\section{IB-INN Model and Loss} 
The model architecture and loss function used in this work were proposed by \cite{ibinn}. 
The loss was derived from an information bottleneck formulation with a hyperparameter, $\beta$, that allows trading off generative and classification capabilities.
The loss function is based on mutual information $I$:
\begin{equation}
    \label{eq:LIB}
    \mathcal{L}_{IB} = I(X, Z) - \beta I(Z, Y). 
\end{equation}
Mutual information quantifies the amount of information which is shared between variables.\footnote{For an invertible mapping $f$ and $Z = f(X)$, $\mathcal{L}_{IB}$ is, in fact, ill-defined, and the authors \cite{ibinn} add noise to $X$ to overcome the issue.}
As such, by minimizing $\mathcal{L}_{IB}$, the mutual information between the input and the latent vector is minimized while the mutual information between the latent vector and class label is maximized.
In practice, the first term, $I(X, Z)$, can be thought of as a generative loss, which results in a good performance on generating images.
The second term, $I(Z, Y)$, is closely related to the categorical cross-entropy loss, thus promoting high accuracy.
Throughout our experiments, we use models trained with the IB-INN loss, $\mathcal{L}_{IB}$.

\begin{figure*}
    \centering
    \includegraphics[width=0.9\textwidth]{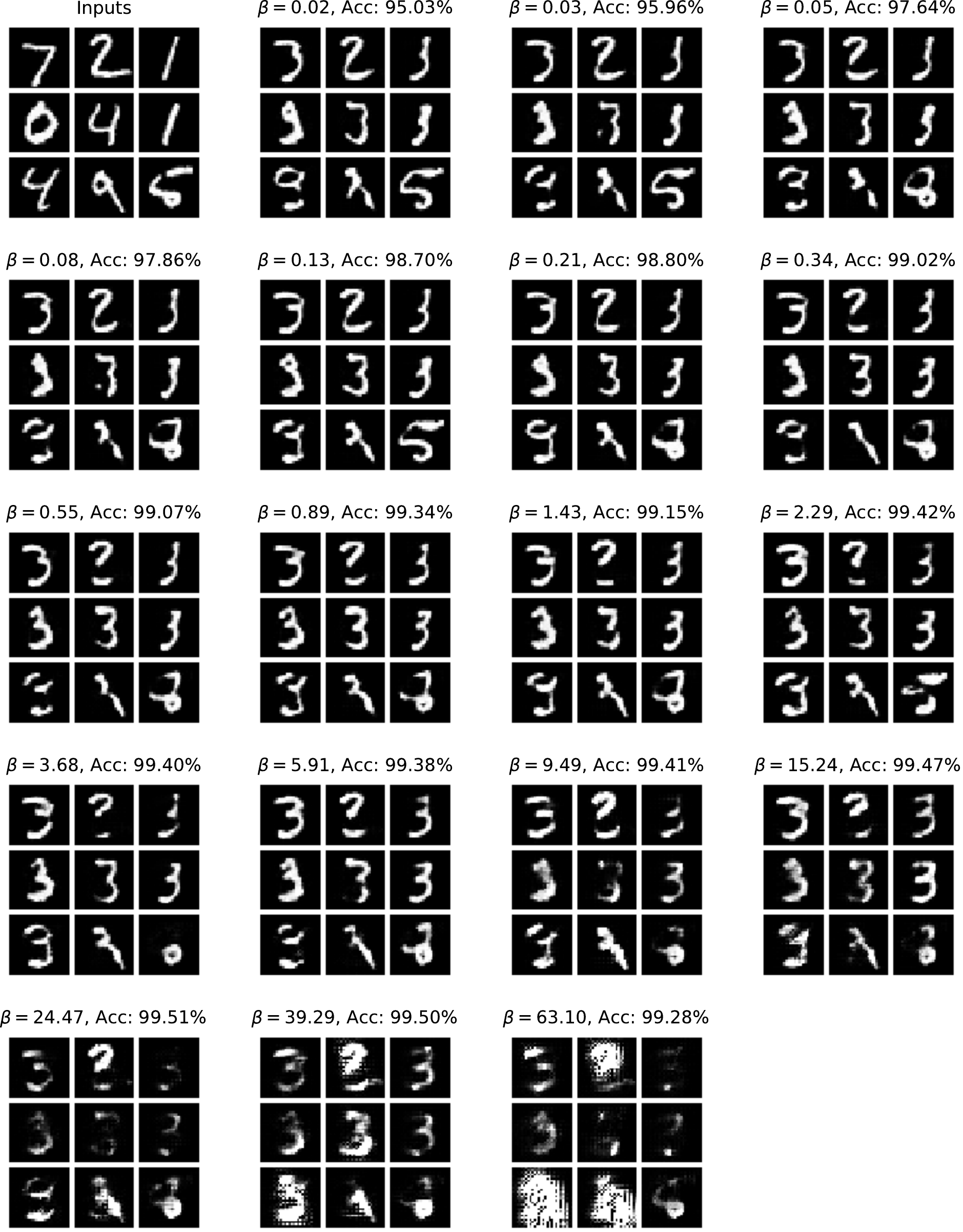}
    \caption{Counterfactual examples for MNIST models trained with different values of $\beta$. The top left square represents the input images that are all changed with target $q=3$. Above plots are $\beta$-values in ascending order and corresponding test set accuracies.}
    \label{fig:betas}
\end{figure*}

For simplicity, we do not conduct experiments across multiple values of $\beta$.
Overall, we find that values close to one strike a good balance between counterfactual examples and model accuracy in our experiments. 
We do, however, include \Fref{fig:betas} which demonstrates the conflicting effect of $\beta$ on the quality of counterfactuals and the accuracy of the model.

\section{Additional Samples}
We include pdfs with extra samples of all figures from the experiments. For each figure, there is a corresponding pdf in the related work zip-file. 
For example, \Fref{fig:fakemnist-samples} has a corresponding pdf in the supplementary material named \verb!figure3.pdf! with additional samples.

\end{document}